\documentclass[10pt,twocolumn,letterpaper]{article}

\usepackage{iccv}
\usepackage{times}
\usepackage{epsfig}
\usepackage{graphicx}
\usepackage{amsmath}
\usepackage{amssymb}

\usepackage{booktabs} 
\usepackage{algorithm}
\usepackage{algorithmic}
\usepackage{subfigure}
\usepackage{url}
\usepackage{multirow}
\usepackage[usenames,dvipsnames,table]{xcolor}

\usepackage[pagebackref=true,breaklinks=true,letterpaper=true,colorlinks,bookmarks=false]{hyperref}

\iccvfinalcopy 

\ificcvfinal\fi
\begin{document}
	
	\title{Computational Baby Learning}
	
	
		\author{Xiaodan~Liang$^{\dagger}$ $^{\star}$ \quad Si~Liu$^{\dagger}$ \quad Yunchao~Wei$^{\dagger}$ $^{\ast}$ \quad Luoqi~Liu$^{\dagger}$ \quad Liang~Lin$^{\star}$ \quad Shuicheng~Yan $^{\dagger}$\\
			$^{\dagger}$ National University of Singapore \quad $^{\star}$ Sun Yat-sen University \quad $^{\ast}$ Beijing Jiaotong university\\
			{\tt\small \{xdliang328,fifthzombiesi,wychao1987,llq667\}@gmail.com}\\
			{\tt\small linliang@ieee.org \quad eleyans@nus.edu.sg}
		}
	\maketitle




\begin{abstract}

Intuitive observations show that a baby may inherently possess the capability of recognizing a new visual concept (e.g., chair, dog) by learning from only very few positive instances taught by parent(s) or others, and this recognition capability can be gradually further improved by exploring and/or interacting with the real instances in the physical world. Inspired by these observations, we propose a computational model for slightly-supervised object detection, based on prior knowledge modelling, exemplar learning and learning with video contexts. The prior knowledge is modeled with a pre-trained Convolutional Neural Network (CNN). When very few instances of a new concept are given, an initial concept detector is built by exemplar learning over the deep features from the pre-trained CNN. Simulating the baby's interaction with physical world, the well-designed tracking solution is then used to discover more diverse instances from the massive online unlabeled videos. Once a positive instance is detected/identified with high score in each video, more variable instances possibly from different view-angles and/or different distances are tracked and accumulated.  Then the concept detector can be fine-tuned based on these new instances. This process can be repeated again and again till we obtain a very mature concept detector. Extensive experiments on Pascal VOC-07/10/12 object detection datasets~\cite{pascalVOCdata} well demonstrate the effectiveness of our framework. It can beat the state-of-the-art full-training based performances by learning from very few samples for each object category, along with about 20,000 unlabeled videos.

\end{abstract}

\section{Introduction}
\vspace{-3mm}

Empirically, we may have the following intuitive observations on how a baby learns\footnote{Note that it does not necessarily mean baby truly learns in this way from neuron-science perspective.}: after the parent(s) or others teach the baby a few instances about a new concept, the initial recognition capability about the concept can be built. During continuously exploring and/or interacting with diverse instances and scenes in real life, the baby can associate the initial simple instances with other variants by using various information linkages. Based on the accumulated instances about the concept, the baby can gradually improve its recognition capability and recognize diverse instances he/she never saw. 

\begin{figure*}[t]
	\begin{center}
		\includegraphics[scale=0.65]{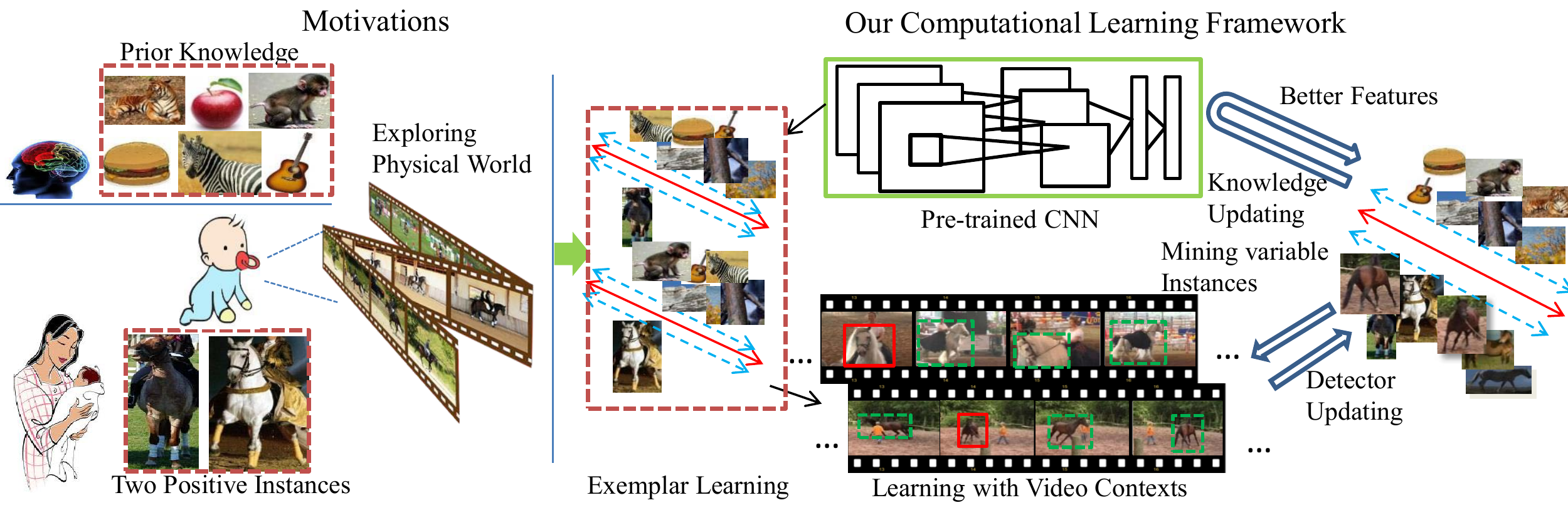}
		\vspace{-2mm}
		\caption{{Illustration of our computational baby learning framework. Inspired by the baby learning process, we integrate prior knowledge modelling, exemplar learning, learning with video contexts for slightly-supervised object detection. The prior knowledge (i.e. feature representation) is modelled with a pre-trained CNN. When very few positive instances of a new concept (e.g., horse) are given, an initial concept detector can be built by exemplar learning based on the feature representation from CNN. Once a positive instance in a frame is detected with the highest score, more variable instances (green dashed box) can be tracked by harnessing video contexts. The concept detector can be gradually improved with these new instances and we repeat this process again and again. In addition, the pre-trained CNN will be gradually fine-tuned if enough instances are collected, which leads to more informative features for training detectors.}}
		\label{fig:framework}
	\end{center}
\vspace{-8mm}
\end{figure*}

Recent successes in computer vision~\cite{googleLeNet}, however, largely rely on the large number of labeled instances of visual concepts, which may require considerable human efforts. The construction of an appearance-based object detector is costly and difficult because the number of training examples must be large enough to capture different variations in the object appearance. Some researchers have made efforts on improving the initial models by using very few labeled data, along with the detection/search results from web images~\cite{NEILiccv13}~\cite{LEARN}~\cite{chen2014enriching} or weakly annotated videos~\cite{2005semiDetection}~\cite{unlabeledvideo13}. 
In this paper, we make the first attempt and build a computational model for slightly-supervised object detection by drawing inspiration from the baby learning process. As illustrated in Figure~\ref{fig:framework}, we propose a robust learning framework which can effectively model the prior knowledge, build the initial model by exemplar learning with very few positive instances for a new concept, and gradually learn a mature object detector by exploring more diverse instances in real-world unlabeled videos. 

First, we model the prior knowledge (i.e. feature representation) with a pre-trained Convolutional Neural network (CNN) in two steps. We first train a generic CNN by the large image classification dataset. The learned convolutional layers provide the effective feature representations for object recognition. We then fine-tune the CNN with the instances of previously learned visual concepts for the domain adaption from object classification to the detection task. Second, when very few positive instances of a new concept are given, the initial concept detector is built by exemplar learning~\cite{malisiewicz2011ensemble}, which trains a separate linear classifier for every exemplar in the training set based on the deep features from intermediate layers of the pre-trained CNN. Other learned visual concepts are used as negative instances to enhance the discriminative capability. Third, we accumulate more variable instances by exploring the massive unlabeled video clips from the online video sharing websites (e.g., YouTube.com). The positive instance with highest confidence in each clip is selected as the seed, and then region-based video tracking is performed to accumulate the variable instances by constraining the appearance consistency and spatial correspondence. The concept detector can thus be progressively improved based on these newly tracked instances. After this process repeats again and again, a very mature concept detector can be obtained. With enough instances for the new concept, the pre-trained CNN can also be further improved/fine-tuned, which can provide better deep features for learning concept detectors. Our framework can thus effortlessly improve a new concept detector based on very few positive instances and large easily-obtained video data. The new concept detector is gradually improved in a never ending way as long as more videos are continuously explored.

Extensive experiments on three challenging object detection datasets (Pascal VOC 07/10/12) well demonstrate the superiority of our computational baby learning framework over other state-of-the-arts~\cite{girshick2013rich}~\cite{ren2013histograms}~\cite{wang2013regionlets}~\cite{fidler2013bottom}. For all three datasets, we only need to learn one detector for each concept, while all previous works train different models for different datasets. Our framework beats other state-of-the-arts by learning from very few positive instances along with about 20,000 unlabeled videos for each object category. 

The contributions of this paper can be summarized as the followings. 1) To the best of our
 knowledge, the proposed framework makes the first attempt to build an effective computational framework for slightly-supervised object detection with inspiration from the baby learning process, where the prior knowledge modelling, exemplar learning and learning with video contexts are integrated. 2)~Only two positive instances are required for learning a new concept detector and then the detector is refined with new variable instances from fully unlabeled videos. There is no assumption that a video must contain a specific object, which makes our framework scalable and robust for learning concept detectors in an online way. 3)~The knowledge of learned concepts can be effectively retained in our model and conveniently utilized to learn new concepts. 

\vspace{-3mm}
\section{Related Work}
\vspace{-3mm}

\textbf{Supervised Learning.} Recently, Convolutional Neural Networks (CNNs) have been shown to perform well in a variety of vision tasks with millions of annotated training images and thousands of categories, including classification~\cite{googleLeNet},  detection~\cite{girshick2013rich} and segmentation~\cite{farabet2013sceneparsing}. Notably, Krizhevsky \emph{et al.}~\cite{krizhevsky2012imagenet} and Szegedy \emph{et al.}~\cite{googleLeNet} achieved great progress in the classification task with large and deep supervised CNN training. Girshick \emph{et al.}~\cite{girshick2013rich} proposed to fine-tune the pre-trained Krizhevsky's network with the PASCAL VOC dataset and achieved the state-of-the-art object detection performance. However, the large performance increase achieved by these methods is only possible due to massive efforts on manually annotating millions of images which can provide good coverage of the space of possible appearance variations. 

\textbf{Semi-Supervised Learning.} To minimize human efforts, some attempts have been devoted to learning reliable models with very few labeled data. Those methods can be summarized into two categories: learning from unlabeled web images (image-based) or video data (video-based). For the first category, existing image-based approaches~\cite{NEILiccv13}~\cite{LEARN} iteratively used image search and detection results to cover more variations. Also text-based~\cite{LEARN} and semantic relationships~\cite{NEILiccv13} were further used to provide more constraints on selecting instances. One problem with these approaches is that the data variations (e.g., different viewpoints or background clutters) cannot be effectively expanded when only with image-based visual similarities. Some other works proposed to transfer the annotated image-level labels~\cite{semisupervisedIm} or ground-truth bounding boxes~\cite{vezhnevets2013associative} from labeled images to unlabeled images for semantically related classes. However, still a lot of labeled images are required to build the adequate subspaces for knowledge transferring.  For the second category, video-based approaches~\cite{weakannvideoDetect12}~\cite{yang2013semi}~\cite{videodiscover} utilized motion cues and appearance correlations within temporal adjacent frames to augment the model training. For example, ~\cite{weakannvideoDetect12} used videos with one class label while our method utilizes many unlabeled videos and very few seed instances. Yang \emph{et al.}~\cite{yang2013semi} used the pre-trained object detector to detect confident or hard samples. Different from~\cite{yang2013semi}, our method investigates how to utilize the video contexts to mine more informative instances. Our slightly-supervised learning means that extremely scare annotated samples (e.g., one or two samples) are used, which is a special case of semi-supervised learning. 
 

\textbf{One-shot Learning.} Our learning framework is partially similar to the one-shot learning~\cite{fei2006one} which learns visual object classifiers by using very few samples. Most of the one-shot learning methods are based on the feature representation transfer~\cite{bart2005cross}, similar geometric context~\cite{hoiem2005geometric} or cross-modal knowledge transfer~\cite{socher2013zero}. However, their performance is far from that of the state-of-the-art object classifiers. By continuously learning from video context, our framework can achieve the state-of-the-art detection results. 

\textbf{Self-paced Learning. } Another related learning pipeline is self-paced learning~\cite{selfPaced2} which learns first from easy samples and then from complex ones. Various vision applications based on self-paced learning have been proposed very recently~\cite{tang2012shifting} ~\cite{selfPacedNips14}. For example, Tang \emph{et al.}~\cite{tang2012shifting} adapted object detectors learned from images to videos by starting with easy samples. Jiang \emph{et al.}~\cite{selfPacedNips14} addressed the data diversity. Our method is a slightly-supervised self-paced learning framework, where very few samples are used as seeds and more instances are iteratively accumulated and learned.

\vspace{-3mm}
\section{Computational Baby Learning Framework}
\vspace{-2mm}


Figure~\ref{fig:framework} shows our proposed framework. Inspired by intuitive observations of the baby learning process, our method integrates prior knowledge modelling, exemplar learning, learning with video contexts for slightly-supervised object detection task. More specifically, the prior knowledge is modeled with a pre-trained CNN. Given very few instances for each new concept, an initial concept detector can be learned with exemplar learning over the deep features from the pre-trained CNN. More difficult instances can be obtained by exploring from real-world unlabelled videos. After that, the detector can be fine-tuned based on these new instances. This process is repeated again and again to obtain a mature detector. The pre-trained CNN can thus be fine-tuned to generate more informative features based on massive mined instances, which may improve the concept detector in turn. 

\vspace{-2mm}
\subsection{Prior Knowledge Modelling}
\vspace{-3mm}
\label{sec:prior}
We model the prior knowledge with two steps. First, we pre-train a general CNN on the ImageNet~\cite{deng2009imagenet} with image-level annotations. In this work, we explore two CNN architectures  for pre-training: the 7-layer architecture by Krizhevsky \emph{et al.}~\cite{krizhevsky2012imagenet} and the Network in Network (NIN) proposed by Lin \emph{et al.}~\cite{imagenet14}~\cite{NIN}. We use the same parameter settings for these two network architectures as in ~\cite{krizhevsky2012imagenet} and~\cite{NIN}. Second, we fine-tune the previous pre-trained CNN with the previously learned concepts for the domain adaption from object classification network into detection network. Since we validate our framework on the PASCAL VOC challenge, we thus use the 179 object classes on the ILSVRC2013 detection dataset as the learned concepts, which excludes the corresponding 21 classes related with the VOC 20 classes. During fine-tuning, we only replace the 1000-way classification layer of the pre-trained CNN with a randomly initialized (N+1)-way classification layer, where N is the number of learned concepts, plus one for background. In our setting, N = 179. We use the validation set (20,121 images) in the ILSVRC2013 detection dataset and only the images that contain at least one object of the 179 classes are used. All region proposals with $\geq$ 0.5 intersection-over-union (IoU) overlap with a ground-truth box are regarded as positives for 179 learned concepts and the rest as negatives. Though our framework can use any category-independent region proposal method, we choose the selective search~\cite{selectiveSearch} to enable a controlled comparison with the previous work~\cite{girshick2013rich}. The CNN fine-tuning starts SGD with a learning rate of $0.001$ for both two networks. For the 7-layer architecture~\cite{krizhevsky2012imagenet}, we uniformly sample
32 positive windows (over all classes) and 96 background
windows to construct a mini-batch of size 128. The fine-tuning is run for 70k SGD iterations and takes 9 hours on a single NVIDIA GeForce GTX TITAN GPU. For NIN~\cite{NIN}, a mini-batch of size $80$, consisting of 20 positive windows and 60 background windows, is used. The fine-tuning is run for 150k SGD iterations and takes 16 hours. Alternatively, our framework can also bootstrap any concept detector without any previously learned concepts, which can be simply implemented by eliminating the fine-tuning step with learned concepts. Our experiments also report the performances when with/without finetuning 179 learned concepts.


\vspace{-3mm}
\subsection{Exemplar Learning}
\vspace{-2mm}

The learned convolution layers in the above pre-trained CNN form the basic feature representation, and then an initial concept detector can be learned based on these deep features and very few positive instances of a new concept.

\textbf{Feature extraction.} 
For all positive and negative instances, we enlarge the tight bounding boxes to contain 16 pixels of image contexts and then wrap it into a fixed 227 $\times$ 227 size as used in~\cite{girshick2013rich}. Deep features are computed as the outputs from the penultimate fully-connected layer (4096-dimension) by forward propagating a mean-subtracted 227 $\times$ 227 image through the pre-trained CNN.

\textbf{Selection of seed instances.}  Our selection strategy of the seeds (including the number of seeds) is optional and our framework can be bootstrapped with any number of seeds of the new concept. For most of our results, we select two common positive instances for each concept from the PASCAL VOC 2007 training set. Specifically, we cluster all positive instances into 10 clusters by k-means. For the top-2 largest clusters, the nearest two samples to the two centroids are selected as the seeds for each concept. Different seed initialization may lead to different model updating and different mined instances in each iteration. However, due to the large number of unlabeled videos, the final detection results can be robust to different seed instances of the same number. We report extensive experiments on the different seed selections (i.e. different seed number and random many times for obtaining different seeds) in Table~\ref{tab:initial}. 

\begin{figure}[t]
	\begin{center}
		\includegraphics[scale=0.33]{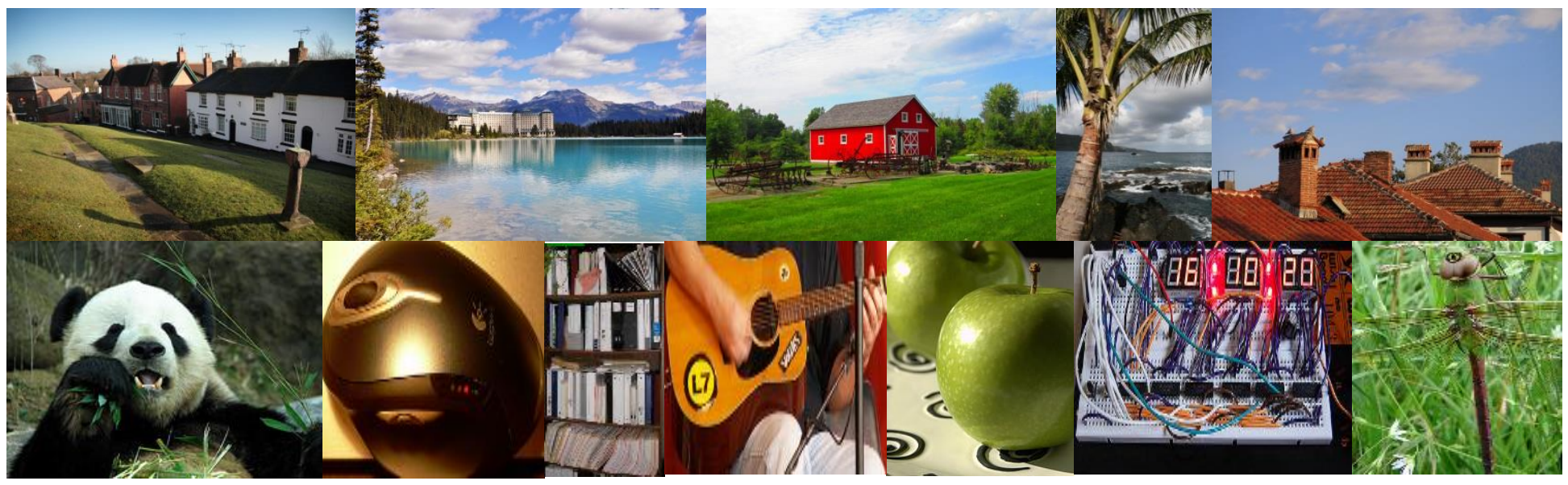}
		\vspace{-1mm}
		\caption{{Some exemplar negative samples. Top row shows the collected general background images and bottom row shows the exemplar instances of previously learned concepts.}}
		\label{fig:negative}
	\end{center}
	\vspace{-11mm}
\end{figure}

\textbf{Negative set collection.} To fairly justify our method, we do not use any annotations of PASCAL VOC Challenge to obtain the negative instances. The negative set used in our framework contains a batch of general background images (i.e., no specific object is included) and learned concept instances. As illustrated in the top row of Figure~\ref{fig:negative}, we collect $4,000$ general scene images from Flickr and use the categories in the SUN scene dataset~\cite{SUNscene} as the search keywords. All region proposals in these background images are used as negative samples. For the learned concepts, the region proposals with $\geq$ 0.5 IoU overlap with the bounding box of $179$ object class instances in the ILSVRC 2013 detection validation set are also treated as negative samples, as shown in the bottom row of Figure~\ref{fig:negative}. Our initial experiment indicates that using general background images, versus our negative set, can result in about 4\% drop in mAP. The possible reason may be that more hard negative samples are included in other object-level concept instances. After more instances of new concepts are collected, our negative set will be gradually enlarged.

\textbf{Exemplar SVM training.} Inspired by~\cite{malisiewicz2011ensemble}, we train a separate linear SVM classifier for each positive instance, and each SVM classifier is highly tuned according to the exemplar's appearance. The exemplar's decision boundary is thus decided, in large part, by the negative samples. For each test image, we thus independently run each exemplar detector and use the non-maximum suppression to create a final set of detections. 

\vspace{-2mm}
\subsection{Learning with Video Contexts}
\label{sec:videocontext}
\vspace{-2mm}
We iteratively improve the concept detectors by mining more variable instances from unlabeled videos. About 20,000 videos for each new concept are crawled from YouTube. Due to the computational limitation, we use the keywords from the VOC dataset collection to prune the videos unrelated to the new concept. The collected video set includes approximate $30\%$ ``noisy" videos that contain none of  the instances of the concept. No manual annotation is performed. In each iteration, we select one seed instance, and region-based tracking is then performed to accumulate the variable instances. The detector and knowledge updating are then performed.

\textbf{Seed instance selection.} In each video clip, there is much redundant information with few appearance differences in temporal adjacent frames. To guarantee appearance variance of tracked instances and limit computational complexity, only key frames of each video are analyzed. We select the image with $\ell_2$ norm of the global GIST~\cite{GIST} feature difference larger than 0.01 as a key frame, compared with its temporal adjacent frames. For all key frames, we perform object detection with the initial detector. We only select the video containing the region with detection score larger than $1$, and the region with the highest score in this video is selected as the seed positive instance.

\begin{figure}[t]
	\begin{center}
		\includegraphics[scale=0.45]{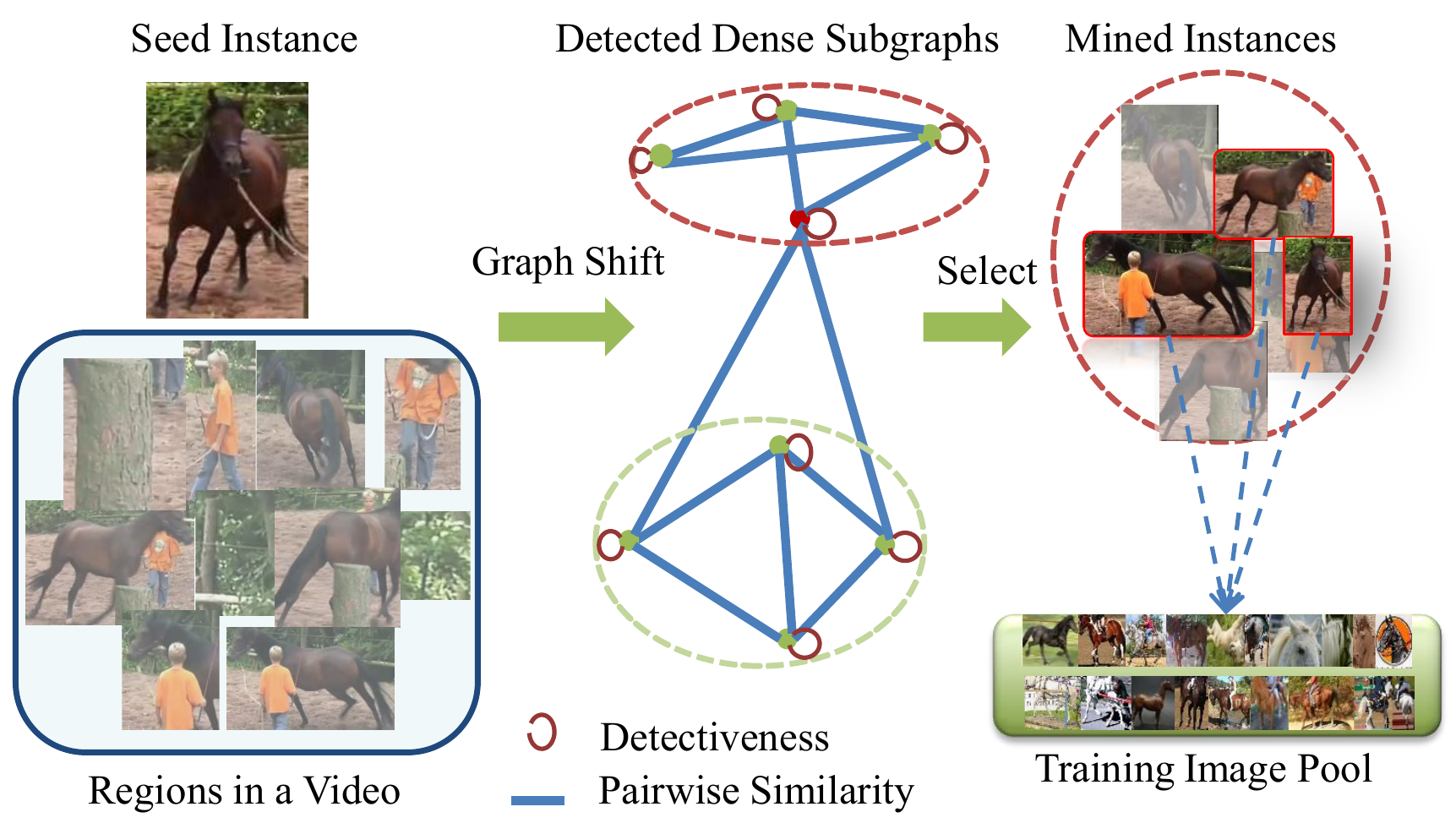}
		\vspace{-2mm}
		\caption{{Region-based video tracking. Given the seed instance, we track other reliable instances from other regions. The affinity graph is built by combining both the pair-wise similarity and the detectiveness of each region. Then dense subgraphs are detected within the affinity graph by graph shift. The subgraph containing the seed instance (red point) is selected. Two instances with top-2 largest similarities with seed instances are placed into the training image pool for fine-tuning the detector in next iteration.}}
		\label{fig:graphsift}
	\end{center}
\vspace{-9mm}
\end{figure}


\textbf{Region-based video tracking.} The region-based video tracking is performed on the selected videos and initialized with their seed instances as illustrated in Figure~\ref{fig:graphsift}. In our framework, we treat the tracking task as a region-based cluster mining problem for both moving and static concepts. Specifically, we extract a batch of region proposals in all key frames using selective search~\cite{selectiveSearch} and represent each region $r_i$ with both the deep feature $\mathbf{x_i}$ and the spatial coordinates $\mathbf{p_i} = (c^x_i, c^y_i, w_i, h_i)$ corresponding to the position, width and height. Since we wish to select the instances from different frames, which may capture more diverse visual patterns, the similarity of two regions from the same frame is thus set as zero. The similarity $A_{i,j}$ for each pair $(r_i, r_j)$ from different frames is thus defined by fusing the appearance similarity and the localization similarity,

\vspace{-6mm}
\begin{equation}
A_{i,j} = \exp\{\frac{||\mathbf{x_i}-\mathbf{x_j}||^2}{\delta_1^2}\} + \alpha(\exp\{\frac{||\mathbf{p_i}-\mathbf{p_j}||^2}{\delta_2^2}\}),
\end{equation}
\vspace{-4mm}

\noindent{where} $\delta_1$ and $\delta_2$ are the empirical variances of $\mathbf{x}$ and $\mathbf{p}$, respectively, and we set $\alpha = 0.3$ because the appearance similarity is more important considering the camera moving and static objects. In addition, to make our framework robust to outliers (i.e., noisy regions), we also constrain the detection score to enlarge the possibility of the region to  contain the concept. We thus define the detectiveness $O_i$ of each region $r_i$ by thresholding the detection score, where $O_i$ of the region with detection score larger than $-3$ is set as 1, otherwise as 0. Note that we do not directly use the detector scores because these variable instances may not be detected by the current detector but can bring more data diversity for further improving the detectors.

To seamlessly integrate the detectiveness of the region and the pairwise similarity, we use the graph shift algorithm~\cite{liu2010robust} to obtain the variable instances, which is efficient and robust for graph mode seeking. This algorithm is particularly suitable for our task as it directly operates on the affinity graph and leaves the outlier points ungrouped. Formally, we define an individual graph $G = (R,A)$ for each video. $R = \{r_1,\dots,r_n\}$ represents all the regions and $A$ is a symmetric matrix with non-negative elements. The diagonal elements of $A$ represent the detectiveness of the regions while the non-diagonal elements measure the pair-wise similarities between regions. The modes of a graph $G$ are defined as local maximizers of the graph density function $g(y) = y^TAy, y\in \Delta^n$, where $\Delta^n = \{y\in R^n: y\geq 0 ~\text{and}~||y||_1 = 1\}$. The strongly connected subgraphs correspond to large local maxima of $g(y)$ over simplex which is an approximate measure of the average affinity score of these subgraphs. And these subgraphs can be found by solving the quadratic optimization problem, i.e., $\max g(y) = y^TAy, y\in \Delta^n$, as in~\cite{liu2010robust}.  The graph
shift algorithm starts from an individual sample and evolves
towards the mode of $G$. The instances reaching the same mode are grouped as a cluster. We can thus select the target subgraph that contains the seed instance. To prevent the rapid semantic drift, we only select two instances in this subgraph, which appear in different frames and have highest similarities with the seed instance. We can thus accumulate much more instances to improve detectors iteratively.

\textbf{Detector Updating.} After accumulating more instances from unlabeled videos, a large set of positive instances of the new concept is collected, which can help improve the concept detector in the next iteration. The frames selected in the previous iterations will not be considered in later iterations, which makes the model equipped with different instances in every iteration. In order to update the concept detector, these newly mined instances are added into the positive set. The regions from general background images and learned concepts are treated as negatives. We retrain one linear SVM classifier for each new concept and the hard negative mining method is also used. After this process repeats again and again, we can achieve a very mature concept detector with a considerable number of mined instances. For fair comparison, we use the same detection strategies as the previous work~\cite{girshick2013rich} in testing phase.

\begin{table*}[htbp]\setlength{\tabcolsep}{3.5pt}
	\centering\scriptsize
	\caption{Detection average precision (\%) on PASCAL VOC2007 test. Rows 1-4 show the baselines. Rows 5-6 are the results of R-CNN based on the NIN~\cite{NIN}. Rows 7-8 show R-CNN results fine-tuned with 179 extra classes. Rows 9-13 show our results in different iterations, with/without fine-tuning and bounding box regression. ``B\_initial'' and ``B\_I15'' represent the results in the beginning with two seeds and after the 15th iteration, respectively. ``B\_FT'' and ``B\_FT\_I2'' are the results after fine-tuning with the mined instances and running 2 more iterations, respectively. ``B\_FT\_I2\_no179\_BB" represents the results after directly using the classification network, and no fine-tuning with 179 detection classes is performed. Rows 15-18 show the results based on NIN~\cite{NIN}.}
	
	\label{tab:mAP07}
	
	\centering
	{
		\begin{tabular}{c|cccccccccccccccccccc|c}
			\toprule
			VOC 2007 test & aero & bike & bird & boat & bottle & bus & car & cat & chair & cow & table & dog & horse & mbike & person & plant & sheep & sofa & train & tv & mAP\\
			\hline
			DPM HSC\cite{ren2013histograms} & 32.2 & 58.3 & 11.5 & 16.3 & 30.6 & 49.9 & 54.8 & 23.5 & 21.5 & 27.7 & 34.0 & 13.7 & 58.1 & 51.6 & 39.9 & 12.4 & 23.5 & 34.4 & 47.4 & 45.2 & 34.3\\ 
			R-CNN\cite{girshick2013rich} & 64.2 & 69.7 & 50.0 & 41.9 & 32.0 & 62.6 & 71.0 & 60.7 & 32.7 & 58.5 & 46.5 & 56.1 & 60.6 & 66.8 & 54.2 & 31.5 & 52.8 & 48.9 & 57.9 & 64.7 & 54.2\\
			R-CNN BB\cite{girshick2013rich} & 68.1 & 72.8 & 56.8 & 43.0 & 36.8 & 66.3 & 74.2 & 67.6 & 34.4 & 63.5 & 54.5 & 61.2 & 69.1 & 68.6 & 58.7 & 33.4 & 62.9 & 51.1 & 62.5 & 64.8 & 58.5\\
			Ngrams \cite{LEARN} & 14.0 & 36.2 & 12.5 & 10.3 & 9.2 & 35.0 & 35.9 & 8.4 & 10.0 & 17.5 & 6.5 & 12.9 & 30.6 & 27.5 & 6.0 & 1.5 & 18.8 & 10.3 & 23.5 & 16.4 & 17.2\\
			\hline	
			R-CNN\_NIN & 72.2 & 74.5 & 58.6 & 47.4 & 38.7 & 68.1 & 75.4 & 72.1 & 38.3 & 69.9 & 57.2 & 69.5 & 67.5 & 72.4 & 59.4 & 34.8 & 61.5 & 60.3 & 67.4 & 69.9 & 61.8\\
			R-CNN\_NIN BB & 72.1 & \textbf{78.2} & 64.3 & \textbf{49.8} & 42.2 & 71.6 & 77.1 & 77.8 & 41.7 & 72.7 & 61.3 & 73.6 & 77.3 & 73.6 & \textbf{64.2} & 37.2 & 64.9 & \textbf{64.5} & \textbf{70.2} & \textbf{72.8} & 65.4\\
			\hline
			R-CNN 179 & 62.5 & 70.2 & 54.4 & 42.7 & 35.4 & 63.1 & 71.9 & 61.5 & 34.0 & 61.0 & 47.1 & 60.7 & 64.1 & 67.9 & 56.8 & 32.6 & 58.2 & 45.7 & 59.2 & 64.5 & 55.7\\
			R-CNN 179 BB & 69.8 & 73.2 & 60.2 & 43.8 & 38.7 & 66.2 & 75.2 & 65.3 & 36.1 & 66.8 & 56.1 & 65.0 & 70.7 & 70.8 & 60.6 & 33.7 & 64.2 & 49.1 & 64.2 & 65.2 & 59.7\\
			\hline
			B\_Initial & 26.3 & 11.9 & 3.2 & 12.9 & 9.3 & 16.0 & 2.5 & 6.4 & 0.9 & 14.3 & 4.1 & 9.7 & 13.2 & 21.6 & 13.7 & 6.0 & 15.9 & 3.5 & 11.1 & 31.4 & 11.7\\
			B\_I15 & 61.1 & 65.7 & 51.1 & 38.8 & 29.8 & 57.4 & 63.8 & 57.8 & 26.8 & 57.1 & 44.3 & 57.2 & 55.7 & 61.3 & 45.5 & 27.2 & 57.1 & 38.0 & 50.4 & 58.0 & 50.3\\
			B\_FT & 65.2 & 71.6 & 53.8 & 39.5 & 32.2 & 64.1 & 70.4 & 63.0 & 33.9 & 60.9 & 50.2 & 58.5 & 64.8 & 65.9 & 54.0 & 27.4 & 60.6 & 45.8 & 59.3 & 60.7 & 55.1\\ 
			B\_FT\_I2 & 68.9 & 70.5 & 55.6 & 42.7 & 37.0 & 64.1 & 71.1 & 66.1 & 34.5 & 63.1 & 51.8 & 60.9 & 63.0 & 67.1 & 52.8 & 31.6 & 62.1 & 45.8 & 57.6 & 64.2 & 56.5\\
			B\_FT\_I2\_BB & 72.2 & 72.8 & 61.8 & 46.7 & 42.0 & 66.1 & 74.2 & 74.6 & 37.3 & 68.3 & 56.8 & 65.7 & 71.3 & 68.4 & 58.0 & 35.1 & 66.3 & 47.2 & 64.0 & 65.7 & 60.7\\
			\hline
			B\_FT\_I2\_no179\_BB & 72.0 & 70.2 & 56.5 & 40.7 & 37.2 & 61.7 & 60.9 & 75.8 & 33.7 & 66.6 & 44.4 & \textbf{76.5} & 69.7 & \textbf{76.9} & 58.5 & 29.3 & 66.9 & 49.7 & 61.1 & 57.6 & 58.3\\
			\hline
			B\_NIN\_I15 & 69.0 & 69.4 & 52.3 & 42.4 & 36.3 & 65.6 & 68.9 & 67.5 & 33.2 & 70.7 & 50.3 & 68.1 & 68.8 & 68.9 & 40.1 & 26.3 & 67.3 & 57.1 & 61.5 & 67.6 & 57.6\\ 
			B\_NIN\_FT & 71.1 & 71.5 & 59.0 & 43.7 & 37.1 & 68.1 & 73.1 & 72.8 & 39.8 & 72.1 & 55.3 & 68.3 & 67.6 & 70.7 & 54.8 & 35.4 & 68.4 & 58.2 & 64.9 & 66.2 & 60.9\\ 
			B\_NIN\_FT\_I2 & 71.0 & 73.6 & 61.3 & 46.3 & 40.6 & 70.3 & 73.8 & 74.0 & 43.7 & 72.9 & 55.6 & 68.5 & 69.2 & 70.7 & 57.6 & 37.7 & 69.3 & 59.6 & 65.3 & 68.7 & 62.5\\ 
			\textbf{B\_NIN\_FT\_I2\_BB} & \textbf{75.9} & {76.8} & \textbf{66.9} & {49.0} & \textbf{47.9} & \textbf{72.1} & \textbf{77.2} & \textbf{77.9} & \textbf{48.6} & \textbf{78.5} & \textbf{65.0} & {73.9} & \textbf{77.3} & {73.6} & {62.7} & \textbf{40.4} & \textbf{73.5} & {64.4} & {69.2} & {70.6} & \textbf{67.1}\\
			\bottomrule
		\end{tabular}
		\vspace{-3mm}
	}
	
\end{table*}

\begin{table*}[htdp]\setlength{\tabcolsep}{3.5pt}
	\caption{{Detection average precision (\%) on PASCAL VOC2010 test.}}\label{tab:mAP10}	
	\centering\scriptsize
	\centering
	{
		\begin{tabular}{c|cccccccccccccccccccc|c}
			\toprule
			VOC 2010 test & aero & bike & bird & boat & bottle & bus & car & cat & chair & cow & table & dog & horse & mbike & person & plant & sheep & sofa & train & tv & mAP\\
			\hline
			Regionlets\cite{wang2013regionlets} & 65.0 & 48.9 & 25.9 & 24.6 & 24.5 & 56.1 & 54.5 & 51.2 & 17.0 & 28.9 & 30.2 & 35.8 & 40.2 & 55.7 & 43.5 & 14.3 & 43.9 & 32.6 & 54.0 & 45.9 & 39.7\\
			SegDPM\cite{fidler2013bottom} & 61.4 & 53.4 & 25.6 & 25.2 & 35.5 & 51.7 & 50.6 & 50.8 & 19.3 & 33.8 & 26.8 & 40.4 & 48.3 & 54.4 & 47.1 & 14.8 & 38.7 & 35.0 & 52.8 & 43.1 & 40.4\\
			R-CNN BB\cite{girshick2013rich} & 71.8 & 65.8 & 53.0 & 36.8 & 35.9 & 59.7 & 60.0 & 69.9 & 27.9 & 50.6 & 41.4 & 70.0 & 62.0 & 69.0 & 58.1 & 29.5 & 59.4 & 39.3 & 61.2 & 52.4 & 53.7\\
			R-CNN 179 BB & 73.1 & 66.2 & 55.2 & 38.7 & 36.5 & 59.2 & 60.2 & 71.8 & 27.1 & 51.6 & 42.0 & 69.5 & 63.9 & 71.0 & 59.2 & 29.7 & 60.0 & 38.5 & 61.8 & 51.2 & 54.3\\
			\hline
			B\_FT\_I2\_BB & 75.7 & 68.0 & 59.2 & 42.6 & 40.0 & 62.4 & 62.0 & 72.3 & 29.5 & 58.2 & 40.8 & 72.0 & 66.3 & 72.8 & 59.9 & 30.9 & 62.6 & 39.9 & 59.0 & 55.7 & 56.5\\
			\textbf{B\_NIN\_FT\_I2\_BB} & \textbf{77.7} & \textbf{73.8} & \textbf{62.3} & \textbf{48.7} & \textbf{45.4} & \textbf{67.3} & \textbf{67.0} & \textbf{80.3} & \textbf{41.3} & \textbf{70.8} & \textbf{49.7} & \textbf{79.5} & \textbf{74.7} & \textbf{78.6} & \textbf{64.5} & \textbf{36.0} & \textbf{69.9} & \textbf{55.7} & \textbf{70.4} & \textbf{61.7} & \textbf{63.8}\\ 			
			\bottomrule
		\end{tabular}
	}
	\vspace{-6mm}
\end{table*}

\begin{table*}[htdp]\setlength{\tabcolsep}{3.5pt}
	\centering\scriptsize
	\caption{{Detection average precision (\%) on PASCAL VOC2012 test.}}\label{tab:mAP12}
	\centering
	{
		
		\begin{tabular}{c|cccccccccccccccccccc|c}
			\toprule
			VOC 2012 test & aero & bike & bird & boat & bottle & bus & car & cat & chair & cow & table & dog & horse & mbike & person & plant & sheep & sofa & train & tv & mAP\\
			\hline
			SDS\cite{hariharan2014simultaneous} & 69.7 & 58.4 & 48.5 & 28.3 & 28.8 & 61.3 & 57.5 & 70.8 & 24.1 & 50.7 & 35.9 & 64.9 & 59.1 & 65.8 & 57.1 & 26.0 & 58.8 & 38.6 & 58.9 & 50.7 & 50.7\\
			R-CNN BB\cite{girshick2013rich} & 71.8 & 65.8 & 52.0 & 34.1 & 32.6 & 59.6 & 60.0 & 69.8 & 27.6 & 52.0 & 41.7 & 69.6 & 61.3 & 68.3 & 57.8 & 29.6 & 57.8 & 40.9 & 59.3 & 54.1 & 53.3\\
			R-CNN\_NIN BB & 77.9 & 73.1 & \textbf{62.6} & 39.5 & \textbf{43.3} & \textbf{69.1} & 66.4 & 78.9 & 39.1 & 68.1 & \textbf{50.0} & 77.2 & 71.3 & 76.1 & \textbf{64.7} & \textbf{38.4} & 66.9 & \textbf{56.2} & 66.9 & \textbf{62.7} & 62.4\\
			R-CNN 179 BB & 72.7 & 66.0 & 55.0 & 34.0 & 32.0 & 59.5 & 60.5 & 70.9 & 29.2 & 51.5 & 40.2 & 70.0 & 62.3 & 68.4 & 58.9 & 30.8 & 58.2 & 37.7 & 58.3 & 53.9 & 53.5\\
			\hline
			B\_FT\_I2\_BB & 75.8 & 68.2 & 58.2 & 39.6 & 37.0 & 63.2 & 62.2 & 72.3 & 29.3 & 59.0 & 40.8 & 71.4 & 66.2 & 71.3 & 59.4 & 30.9 & 61.2 & 41.1 & 57.3 & 56.5 & 56.0\\
			\textbf{B\_NIN\_FT\_I2\_BB} & \textbf{78.0} & \textbf{74.2} & {61.3} & \textbf{45.7} & {42.7} & {68.2} & \textbf{66.8} & \textbf{80.2} & \textbf{40.6} & \textbf{70.0} & {49.8} & \textbf{79.0} & \textbf{74.5} & \textbf{77.9} & {64.0} & {35.3} & \textbf{67.9} & {55.7} & \textbf{68.7} & {62.6} & \textbf{63.2}\\
			\bottomrule
		\end{tabular}
	}
	\vspace{-3mm}		
\end{table*}

\begin{table*}[htbp]\setlength{\tabcolsep}{3.2pt}
	\centering\scriptsize
	\caption{{Detection average precision (\%) on PASCAL VOC2007 test by the version initialized with more training data.  } }
	\label{tab:mAPvoc07}
	
	\centering
	{
		\begin{tabular}{c|cccccccccccccccccccc|c}
			\toprule
			VOC 2007 test & aero & bike & bird & boat & bottle & bus & car & cat & chair & cow & table & dog & horse & mbike & person & plant & sheep & sofa & train & tv & mAP\\
			\hline
			R-CNN BB\cite{girshick2013rich} & 68.1 & 72.8 & 56.8 & 43.0 & 36.8 & 66.3 & 74.2 & 67.6 & 34.4 & 63.5 & 54.5 & 61.2 & 69.1 & 68.6 & 58.7 & 33.4 & 62.9 & 51.1 & 62.5 & 64.8 & 58.5\\
			R-CNN\_NIN BB & 72.1 & \textbf{78.2} & 64.3 & 49.8 & 42.2 & 71.6 & 77.1 & 77.8 & 41.7 & 72.7 & 61.3 & 73.6 & 77.3 & 73.6 & 64.2 & 37.2 & 64.9 & 64.5 & \textbf{70.2} & 72.8 & 65.4\\
			\hline
			{B\_VOC\_I2} & 69.2 & 70.3 & 58.5 & 42.7 & 38.3 & 64.6 & 71.7 & 67.3 & 37.2 & 64.7 & 52.1 & 62.6 & 64.6 & 69.1 & 54.1 & 33.3 & 62.7 & 46.6 & 58.8 & 66.8 & 57.8\\ 
			{B\_VOC\_I2\_BB} & {72.3} & {72.7} & {64.0} & {46.7} & {43.5} & {67.5} & {74.8} & {74.5} & {39.4} & {70.1} & {57.7} & {68.3} & {71.5} & {70.2} & {58.5} & {36.9} & {68.1} & {49.1} & {65.5} & {67.9} & {62.0}\\
			\hline
			{B\_NIN\_VOC\_I2} & 73.6 & 74.8 & 65.5 & 48.4 & 46.4 & 71.2 & 74.6 & 76.0 & 46.1 & 75.3 & 56.8 & 72.6 & 73.6 & 73.8 & 60.1 & 40.5 & 71.6 & 68.3 & 67.4 & 73.1 & 65.5\\
			\textbf{B\_NIN\_VOC\_I2\_BB} &  \textbf{76.2} & {77.8} & \textbf{69.9} & \textbf{51.0} & \textbf{52.0} & \textbf{73.6} & \textbf{77.5} & \textbf{78.0} & \textbf{49.4} & \textbf{82.1} & \textbf{65.2} & \textbf{76.8} & \textbf{79.1} & \textbf{74.9} & \textbf{64.5} & \textbf{41.6} & \textbf{74.7} & \textbf{70.8} & {69.8} & \textbf{73.6} & \textbf{68.9}\\ 
			\bottomrule
		\end{tabular}	
	}
	\vspace{-6mm}		
\end{table*}

\textbf{Knowledge Updating.} Once enough instances of each new concept (about 10,000 instances) are obtained, the pre-trained CNN can be further improved to generate more informative features by fine-tuning it with these new instances. During fine-tuning, we replace the (N+1)-way output layer of the pre-trained CNN in Section~\ref{sec:prior} with a randomly initialized (M+1)-way classification layer (including M new concepts and one for background). We set $M=20$ in our experiments. Since these mined instances may contain some noisy data (e.g., inaccurate bounding box of the object), we only use the original set of mined instances during fine-tuning and no additional data augmentation (e.g., $\geq$ 0.5 IoU overlap) is performed. The negatives for training concept detectors are used as background. The fine-tuning is run for 50K SGD iterations for the 7-lay architecture~\cite{krizhevsky2012imagenet} and 100K iterations for NIN~\cite{NIN}, respectively.  

Finally, a bounding box regression model is learned to fix many localization errors in the testing as in~\cite{girshick2013rich}. From the mined instances, we select $2,000$ detected instances with highest detection scores in the later iterations as ground-truth boxes for training the regression model. The concept detector in the later iterations will be very mature and these top detection boxes have high possibilities to locate the precise object locations. We only consider a region proposal if it has an IoU with ground-truth box greater than $0.8$.

\vspace{-1mm}
\section{Experiments}
\vspace{-1mm}


We evaluate our framework on the PASCAL Visual Object Classes (VOC) datasets~\cite{pascalVOCdata}, which are widely used as the benchmark for object detection. PASCAL VOC 2007, VOC 2010 and VOC 2012 are all tested. For each object class, we train the object detector by using two seeds and about $20,000$ unlabeled videos. Note that our method is independent of any specific test set and only one object detector is used for testing the three datasets. For VOC 2010 and VOC 2012, we evaluate test results on the online evaluation server. We compare our method with the state-of-the-art baselines, including DPM HSC~\cite{ren2013histograms}, Regionlets~\cite{wang2013regionlets}, SegDPM~\cite{fidler2013bottom} and R-CNN~\cite{girshick2013rich}. They used all data in the VOC \emph{train} and \emph{val} set for training detectors. We also compare our method with the recent weakly supervised method~\cite{LEARN}, which discovers instances from web images. We use 179 extra classes in the ILSVRC 2013 detection task to fine-tune the pre-trained CNN by 1000 classes in the ILSVRC 2012 classification. We implement two versions of R-CNN (i.e., ``R-CNN 179'' and ``R-CNN 179 BB'' with bounding-box regression), which firstly fine-tune the classification CNN with 179 extra classes and then fine-tune the CNN with VOC 20 classes following the settings in~\cite{girshick2013rich}. We also report the performances of two version of R-CNN (i.e., ``R-CNN\_NIN" and ``R-CNN\_NIN BB") using the Network-in-Network (NIN)~\cite{NIN}.

\vspace{-2mm}
\subsection{Comparison with the state-of-the-arts}
\vspace{-2mm}

Table~\ref{tab:mAP07}, ~\ref{tab:mAP10} and \ref{tab:mAP12} shows the results on the VOC 2007, 2010 and 2012, respectively. ``R-CNN\_NIN BB" can significantly increase the mAP on VOC 2007 achieved by ~\cite{girshick2013rich} from $58.5\%$ to $65.4\%$ and mAP on VOC 2012 of ~\cite{girshick2013rich} from $53.3\%$ to $62.4\%$, respectively. Its superiority largely benefits from the better neural network architecture. The ``R-CNN 179 BB'' only performs slightly better than ``R-CNN BB'' (e.g., smaller than 0.6\% increase on VOC 2010 and VOC 2012). The main reason is that the samples from 179 extra classes provide limited additional information when enough instances of 20 classes are already used in~\cite{girshick2013rich}. However, when only two instances of a new concept are given, our method can benefit from these instances for domain-specific fine-tuning. All our variants strongly outperform the methods~\cite{ren2013histograms}~\cite{wang2013regionlets}~\cite{fidler2013bottom} based on hand-crafted features learning and deformable part models. Based on the 7-layer network~\cite{krizhevsky2012imagenet}, our method (``B\_FT\_I2\_BB'') achieves $60.7\%$ in mAP, which is significantly superior to $58.5\%$ of ``R-CNN''~\cite{girshick2013rich} and $34.3\%$ of ``DPM HSC''~\cite{girshick2013rich}. Compared to R-CNN, our method increases the performance by $2.8\%$ and $2.7\%$ on VOC2010 and VOC2012, respectively. After fine-tuning the classification network with 179 extra classes, the performance of our method (``B\_FT\_I2\_BB'') can increase by $2.4\%$ over ``B\_FT\_I2\_no179\_BB''. It well verifies the importance of domain adaptation from classification into detection. When fine-tuning the CNN based on the Network in Network (NIN)~\cite{NIN}, our method (``B\_NIN\_FT\_I2\_BB'') can achieve 67.1\% on VOC2007, 63.8\% on VOC 2010, and 63.2\% on VOC2012, which outperforms the ``R-CNN BB~\cite{girshick2013rich}'' by a large margin of more than 8\% on all three test sets and significantly outperforms the ``R-CNN\_NIN BB'' by $1.7\%$ on VOC2007. The bounding box regression can further fix a large number of mislocalized detections. Note that our method only uses two positive instances and trains one single detector for all three datasets, while the baselines use different large training sets and carefully tune the model parameters for different test sets. This superiority well verifies the effectiveness and generality of our framework that automatically learns a significantly better detector than the fully supervised methods. The recent weakly supervised method~\cite{LEARN} only obtained $17.2\%$ in mAP on VOC 2007, which is much worse than the proposed method.

\begin{table*}[htbp]\setlength{\tabcolsep}{3.2pt}
	\centering\scriptsize
	\caption{{Detection average precision (\%) on PASCAL VOC2012 test by the version initialized with more training data.} }
	\label{tab:mAPvoc12}
	
	\centering
	{
		\begin{tabular}{c|cccccccccccccccccccc|c}
			\toprule
			VOC 2012 test & aero & bike & bird & boat & bottle & bus & car & cat & chair & cow & table & dog & horse & mbike & person & plant & sheep & sofa & train & tv & mAP\\
			\hline
			SDS\cite{hariharan2014simultaneous} & 69.7 & 58.4 & 48.5 & 28.3 & 28.8 & 61.3 & 57.5 & 70.8 & 24.1 & 50.7 & 35.9 & 64.9 & 59.1 & 65.8 & 57.1 & 26.0 & 58.8 & 38.6 & 58.9 & 50.7 & 50.7\\
			R-CNN BB\cite{girshick2013rich} & 71.8 & 65.8 & 52.0 & 34.1 & 32.6 & 59.6 & 60.0 & 69.8 & 27.6 & 52.0 & 41.7 & 69.6 & 61.3 & 68.3 & 57.8 & 29.6 & 57.8 & 40.9 & 59.3 & 54.1 & 53.3\\
			R-CNN\_NIN BB & 77.9 & 73.1 & 62.6 & 39.5 & 43.3 & 69.1 & 66.4 & 78.9 & 39.1 & 68.1 & 50.0 & 77.2 & 71.3 & 76.1 & 64.7 & \textbf{38.4} & 66.9 & 56.2 & 66.9 & 62.7 & 62.4\\
			\hline
			{B\_NIN\_VOC\_I2} & 77.6 & 71.7 & 60.9 & 41.3 & 38.2 & 65.5 & 64.5 & 80.0 & 38.1 & 69.9 & 47.1 & 79.3 & 74.7 & 76.1 & 61.9 & 33.5 & 67.7 & 54.8 & 62.6 & 63.2 & 61.4\\ 
			\textbf{B\_NIN\_VOC\_I2\_BB} & \textbf{80.2} & \textbf{75.0} & \textbf{64.9} & \textbf{45.8} & \textbf{44.0} & \textbf{70.1} & \textbf{67.6} & \textbf{81.4} & \textbf{40.8} & \textbf{71.4} & \textbf{51.9} & \textbf{81.0} & \textbf{75.6} & \textbf{78.2} & \textbf{66.1} & {37.6} & \textbf{68.5} & \textbf{59.4} & \textbf{68.0} & \textbf{65.2} & \textbf{64.6}\\ 
			\bottomrule
		\end{tabular}	
	}
	\vspace{-3mm}		
\end{table*}

\begin{table*}[htbp!]\setlength{\tabcolsep}{2.8pt}
	\centering\scriptsize
	\caption{{Detection average precision (\%) on PASCAL VOC2007 test by excluding the VOC classes during pretraining the CNN. } }
	\label{tab:novoc}
	
	\centering
	{
		\begin{tabular}{c|cccccccccccccccccccc|c}
			\toprule
			VOC 2007 test & aero & bike & bird & boat & bottle & bus & car & cat & chair & cow & table & dog & horse & mbike & person & plant & sheep & sofa & train & tv & mAP\\
			\hline
			{B\_NIN\_noVOC\_Initial} & 20.6 & 0.5 & 0.4 & 14.2 & 3.6 & 4.0 & 12.4 & 20.4 & 11.4 & 3.4 & 11.7 & 24.1 & 13.5 & 2.9 & 12.1 & 10.9 & 23.4 & 2.1 & 25.9 & 2.9 & 11.0\\			
			{B\_NIN\_noVOC\_I15} & {65.0} & {71.6} & {53.9} & {39.7} & {31.9} & {64.1} & {70.2} & {62.8} & {34.3} & {61.0} & {50.3} & {59.4} & {65.1} & {66.4} & {54.2} & {28.9} & {61.4} & {45.5} & {59.4} & {61.1} & {55.3}\\ 
			{B\_NIN\_noVOC\_FT} & 70.4 & 73.0 & 59.1 & 42.8 & 36.6 & 66.4 & 74.2 & 69.4 & 36.3 & 66.7 & 57.9 & 65.3 & 70.4 & 66.7 & 56.7 & 32.9 & 65.0 & 53.6 & 63.3 & 64.4 & 59.5\\ 
			{B\_NIN\_noVOC\_FT\_I2} & {69.3} & {72.9} & {58.0} & {44.6} & {40.5} & {67.3} & {74.0} & {72.7} & {39.5} & {71.0} & {56.3} & {66.2} & {67.4} & {73.0} & {58.4} & {34.8} & {64.5} & {60.2} & {65.0} & {67.4} & {61.2}\\
			\textbf{B\_NIN\_noVOC\_FT\_I2\_BB} & \textbf{73.4} & \textbf{76.3} & \textbf{63.1} & \textbf{50.2} & \textbf{44.9} & \textbf{71.3} & \textbf{77.0} & \textbf{75.8} & \textbf{44.1} & \textbf{74.7} & \textbf{64.4} & \textbf{71.9} & \textbf{74.4} & \textbf{74.5} & \textbf{64.0} & \textbf{37.5} & \textbf{67.4} & \textbf{63.6} & \textbf{69.5} & \textbf{68.7} & \textbf{65.3}\\
			\bottomrule
		\end{tabular}
	}
	
\vspace{-4mm}		
\end{table*}

				\begin{figure}
					\begin{center}
						\includegraphics[scale=0.18]{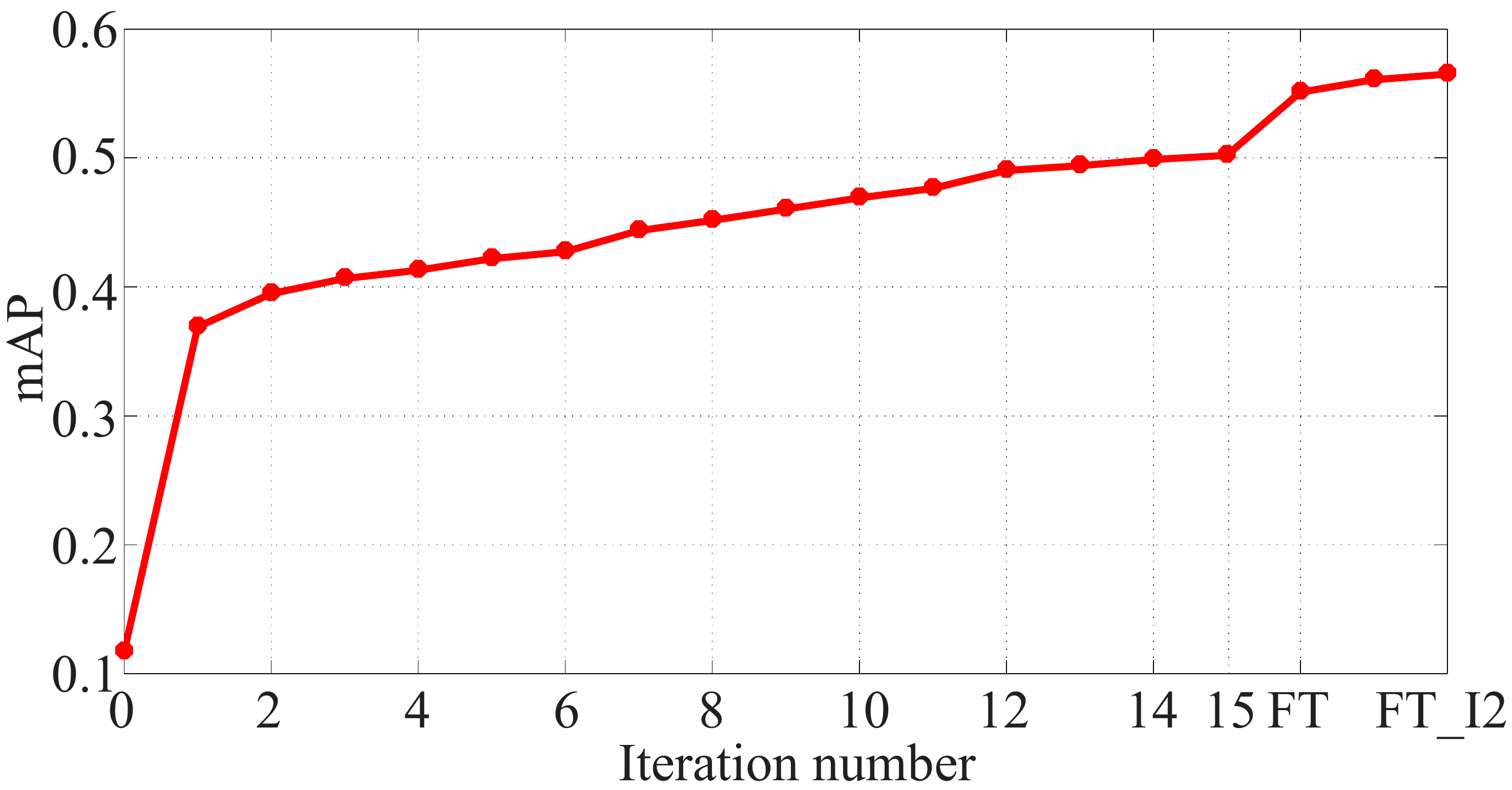}
						\vspace{-1mm}
						\caption{{Performances in different iterations of our framework on VOC 2007. We run 15 iterations for mining more instances. Then we fine-tune the CNN with mined instances (``FT'') and 2 iterations are further performed to improve the detectors (``FT\_I2'').}}
						\label{fig:round}
					\end{center}
					\vspace{-8mm}	
				\end{figure}
				
\vspace{-2mm}
\subsection{Computational Baby Learning Results}
\vspace{-2mm}

Figure~\ref{fig:round} shows our performances in different iterations as well as before/after fine-tuning with more variable instances. We show the results based on the 7-layer network on VOC 2007 and the corresponding AP for each class is presented in Table~\ref{tab:mAP07}. In the beginning, we only obtain 11.7\% in mAP with only two seeds. After the first round of learning with video context, we can substantially improve the mAP to $36.1\%$, which is even higher than mAP of DPM HSC~\cite{ren2013histograms}. Most of the easy test samples can be detected by our updated model. After 15 iterations are performed, we can achieve 50.3\% in mAP (``B\_I15'') and collect about 10,000 instances for each class. We then further fine-tune the pretrained CNN with these mined instances and 4.8\% improvement is achieved (``B\_FT''). Using NIN as the CNN architecture, we achieve $57.6\%$ after 15 iterations (``B\_NIN\_I15'') and obtain $60.9\%$ after fine-tuning (``B\_NIN\_FT''). It proves that more informative features can be learned by fine-tuning. And the superiority of ``B\_NIN\_FT'' over ``B\_FT'' well demonstrates that more informative CNNs can lead to better initialization of our framework and learning capabilities during the repetition of the process. We then further improve the detectors by running 2 more iterations based the fine-tuned CNNs and better detection performances can be achieved, shown by ``B\_FT\_I2'', ``B\_NIN\_FT\_I2''. For both static (e.g., chair) and moving objects (e.g., bicycle), our model can be gradually improved, which speaks well for the effectiveness of our region-based video tracking. Limited by the computational cost, our experiments only report the current results at two iterations after fine-tuning CNN. With better CNN architectures (e.g., googleLeNet~\cite{googleLeNet}), it is predictable that our detectors can be further improved.


\begin{table}\setlength{\tabcolsep}{2.6pt}
	\caption{{Detection average precision (\%) on aeroplane class of PASCAL VOC2007 by different seed selections. We run our version ``B\_FT\_I2" with the same setting by using different numbers of randomized seed instances. We report the results based on different numbers of seeds, i.e., one, two and five, as well as different seed instances randomly ten times for each number of seed.} }
	
	\centering\scriptsize
	\label{tab:initial}\begin{tabular}{c|cccccccccc|cccccccccc|c}
		\toprule
		seed number & 1 & 2 & 3 & 4 & 5 & 6 & 7 & 8 & 9 & 10 & mean \\
		\hline
		1 & 65.1 & 66.1 & 68.1 & 67.3 & 67.2 & 68.2 & 65.4 & 66.1 & 67.8 & 66.7 & 66.8\\
		2 & 67.0 & 66.8 & 70.4  & 69.7 & 68.3 & 69.2 & 66.9 & 67.9 & 69.2 & 69.8 & 68.5\\
		5 & \textbf{69.7} & \textbf{69.3} & \textbf{71.2} & \textbf{72.3} & \textbf{70.1} & \textbf{70.8} & \textbf{71.5} & \textbf{70.9} & \textbf{70.2} & \textbf{71.3} & \textbf{70.7}\\
		\bottomrule
	\end{tabular}
	\vspace{-6mm}
\end{table}

\vspace{-2mm}
\subsection{Discussion on Different Seed Selections}
\vspace{-2mm}

We extensively evaluate how our framework performs with different seed selections. Due to the computational limitation, we only test on one specific object class, i.e., aeroplane, as reported in Table~\ref{tab:initial}. We test three numbers of seeds during the initialization. For each number, we  generate different seeds randomly ten times to evaluate the robustness on seed selections. It can be seen that our method can archive better performance with more initial seeds. With different randomized seeds of each number, we obtain slightly different results and their mean $68.5\%$ is only slightly worse than $68.9\%$ by our version with two selected instances in Table~\ref{tab:mAP07}. The CNN fine-tuning only with the aeroplane class may lead to this slightly decrease. The main reason for the robustness may be the usage of the large number of unlabelled videos. By mining various instances with different views or appearance changes, we can easily introduce greater data diversity into the model training.

\vspace{-2mm}
\subsection{Initialization with More Training Data}
\vspace{-2mm}

We evaluate the state-of-the-arts (e.g., R-CNN) trained with all training sets can be further improved by using our framework. The results of ``B\_VOC\_I2'' and ``B\_VOC\_I2\_BB'' are shown in Table~\ref{tab:mAPvoc07}, in which all the images in VOC 2007 are used. The detectors are trained over the deep features from the 7-layer CNN and two more iterations are performed to mine more instances from videos. We obtain $62.0\%$ mAP on VOC 2007, $3.5\%$ higher than the original ``R-CNN BB'' (58.5\% in mAP). Based on the Network-in-Network, we also achieve superior performances over ``R-CNN\_NIN BB" ($68.9\%$ vs $65.4\%$ in mAP on VOC 2007 and $64.6\%$ vs $62.4\%$ in mAP on VOC 2012 in Table~\ref{tab:mAPvoc07} and Table~\ref{tab:mAPvoc12}, respectively).

\begin{figure}
	\begin{center}
		\includegraphics[scale=0.58]{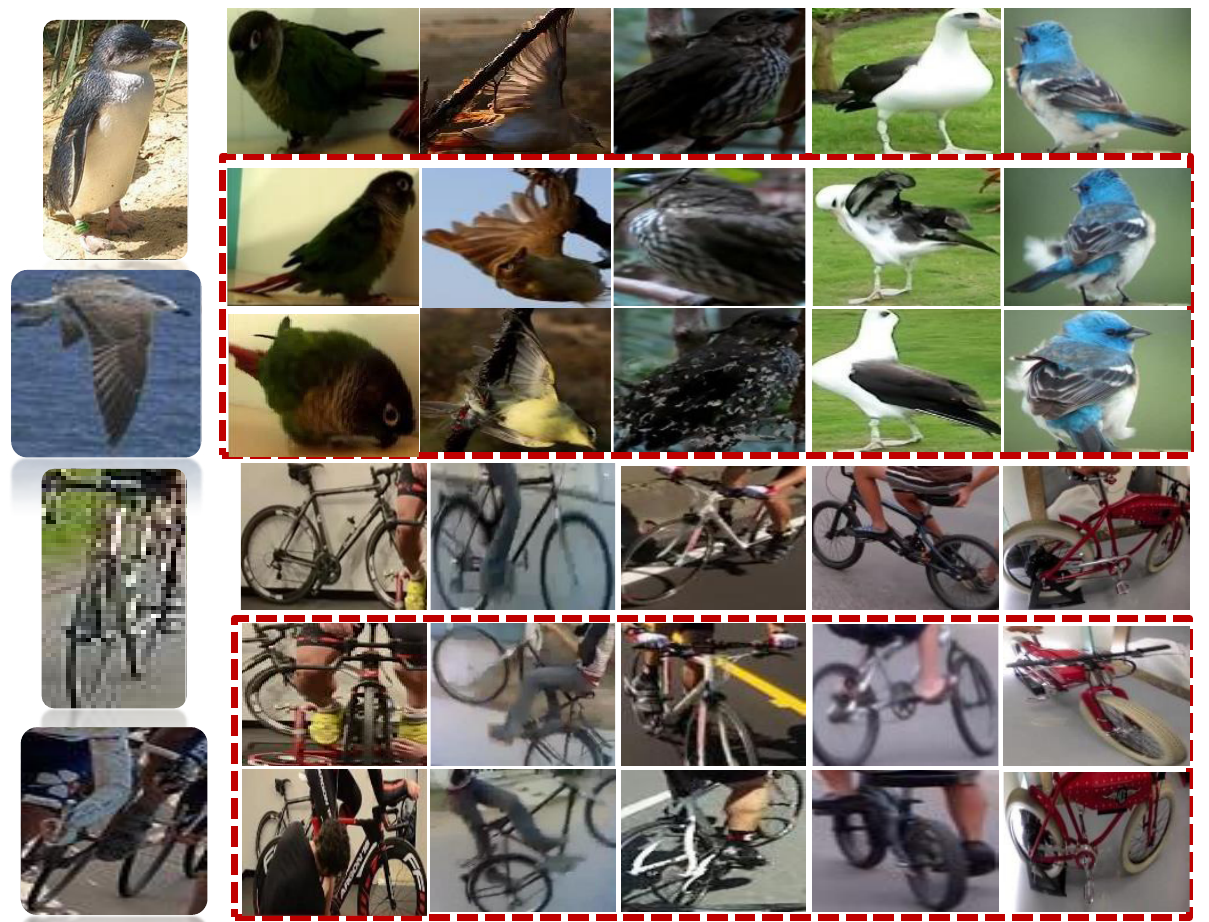}
		\vspace{-1mm}
		\caption{{Visualization of our tracking results for bird and bicycle. For each class, the left column shows the initialized two seeds. The top row shows the detected seeds in each video and two tracked instances are presented within the dashed box.}}
		\label{fig:tracking}
	\end{center}
	\vspace{-8mm}
\end{figure}

\vspace{-2mm}
\subsection{Pretraining Prior Knowledge CNN After Excluding the VOC Classes}
\vspace{-1mm}

Recall that in Section~\ref{sec:prior}, we pre-train the CNN on the ImageNet dataset with 1000 classes. The 1000 classes on ImageNet include similar classes or sub-classes with VOC 20 categories (e.g., space shuttle vs aeroplane and Chihuahua vs dog). To make more strict comparisons and analyses, we conduct experiments by excluding the similar classes with the VOC 20 classes when training the general CNN. We manually exclude the $214$ similar classes with VOC 20 classes of 1000 classes in ImageNet, such as studio couch and day bed vs sofa, laptop computer vs tvmonitor in ImageNet and VOC classes, respectively. Then only $789$ classes in ImageNet are used for pretraining. We train this CNN based on the Network in Network (NIN) and use the same parameter setting  as in~\cite{NIN}. Based on this retrained CNN, we report the corresponding performance on VOC 2007 in Table~\ref{tab:novoc}. Our method can still substantially improve its initial performance with only two seeds. The mAP achieved by ``B\_NIN\_noVOC\_FT\_I2\_BB" is comparable with the fully-trained ``R-CNN\_NIN\_BB" version, i.e., $65.3\%$ vs $65.4\%$ on VOC 2007, respectively. 

\begin{figure}
	\begin{center}
		\includegraphics[scale=0.6]{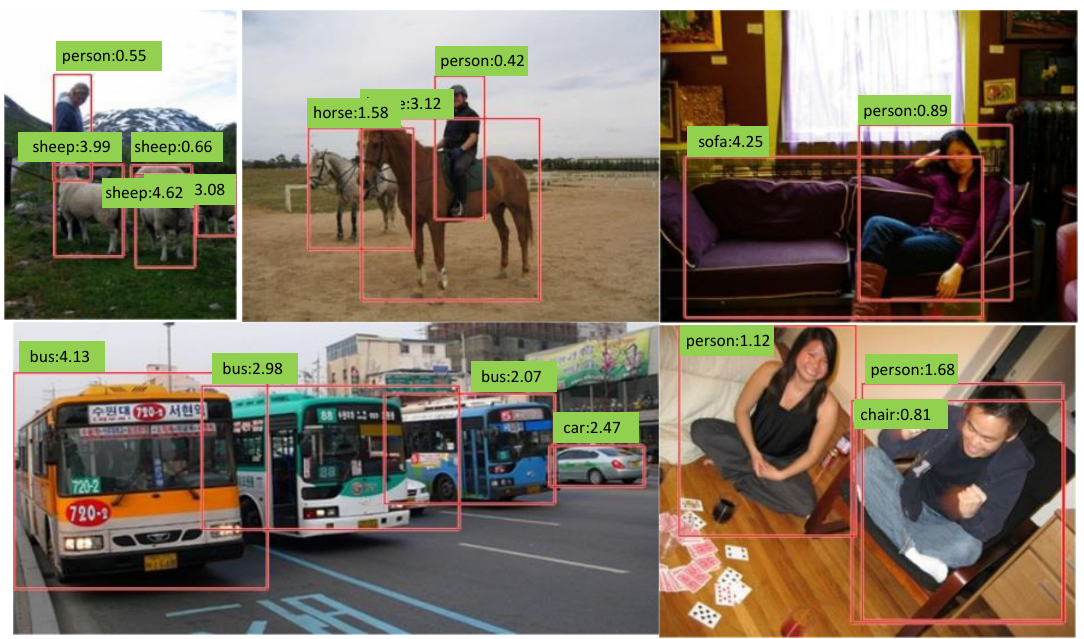}
		\vspace{-1mm}
		\caption{{Some exemplar detection results. All detections with precision greater than 0.5 are shown. Each detection is labeled with the predicted class and the detection score from the detector. View digitally with recommended zoom.}}
		\label{fig:detection}
	\end{center}
\vspace{-8mm}
\end{figure}

\vspace{-2mm}
\subsection{Visualizations}
\vspace{-2mm}

We show the two selected instances and some mined instances for four classes in Figure~\ref{fig:tracking}. We randomly select some mined instances from all iterations. It can be observed that our method successfully tracks variable instances with different view-angles, occlusions or appearance variance. Many qualitative detection results on the VOC 2007 test set are presented in Figure~\ref{fig:detection}, which are obtained from our best model ``B\_NIN\_FT\_I2\_BB'', each image is selected due to it is impressive and accurate. 

\vspace{-3mm}
\section{Conclusion and Future Work}
\vspace{-2mm}

In this paper, inspired by the intuitive observation of the baby learning process, we presented a novel computational slightly-supervised learning framework for object detection by combining prior knowledge modeling, exemplar learning, and learning with video contexts. Significant improvements over fully-training based methods were achieved by our framework on PASCAL VOC 07/10/12 with only two positive instances along with about 20,000 unlabeled real-world videos. In the future, we will explore how to adequately utilize more contextual information (e.g. scene, human actions, other objects) to mine more accurate and diverse instances. Our framework can also be easily extended to improve various vision tasks, such as face age recognition, people identification and scene classification. 

\vspace{-2mm}

{\small
	\bibliographystyle{ieee}
	\bibliography{arxiv}
}

\end{document}